\documentclass[10pt,twocolumn,letterpaper]{article}

\usepackage{iccv}
\usepackage{times}
\usepackage{epsfig}
\usepackage{graphicx}
\usepackage{amsmath}
\DeclareMathOperator*{\argmin}{argmin} 
\DeclareMathOperator*{\argmax}{argmax} 
\usepackage{amssymb}
\usepackage[ruled,linesnumbered]{algorithm2e}
\usepackage{booktabs}
\usepackage{colortbl}
\usepackage{xcolor}
\usepackage{caption}
\usepackage{subfigure}

\usepackage[pagebackref=true,
breaklinks=true,
letterpaper=true,
colorlinks,
citecolor=black!30!green,
bookmarks=false]{hyperref}

\newcommand{\trm} {\textrm}

\newcommand{\tbf} {\textbf}
\newcommand{\sbf} {\boldsymbol}
\newcommand{\vx} {\textbf{\textrm{x}}}
\newcommand{\mcal}{\mathcal}
\newcommand{\ssubsec}{\vspace{0pt}\tbf}
\newcommand{\myrule}{\specialrule{1pt}{0pt}{0pt}}
\newcommand{\mytinyrule}{\specialrule{.1pt}{0pt}{0pt}}
\usepackage[accsupp]{axessibility}

\iccvfinalcopy 

\def\iccvPaperID{2854} 

\begin{document}

\title{ImbSAM: A Closer Look at Sharpness-Aware Minimization in Class-Imbalanced Recognition}

\author{
Yixuan Zhou\textsuperscript{1} \and Yi Qu\textsuperscript{1} \and Xing Xu\textsuperscript{1,}\thanks{Corresponding author.} \and Hengtao Shen\textsuperscript{1,2} \and 
\textsuperscript{1}Center for Future Media \& School of Computer Science and Engineering, \\
University of Electronic Science and Technology of China
\quad \textsuperscript{2}Peng Cheng Laboratory, China \\
{\tt\small yxzhou@std.uestc.edu.cn, iquyiiii@gmail.com, xing.xu@uestc.edu.cn, shenhengtao@hotmail.com} 
}

\maketitle

\begin{abstract}
   Class imbalance is a common challenge in real-world recognition tasks, where the majority of classes have few samples, also known as tail classes.
   We address this challenge with the perspective of generalization and empirically find that the promising Sharpness-Aware Minimization (SAM) fails to address generalization issues under the class-imbalanced setting.
   Through investigating this specific type of task, we identify that its generalization bottleneck primarily lies in the severe overfitting for tail classes with limited training data.
   To overcome this bottleneck, we leverage class priors to restrict the generalization scope of the class-agnostic SAM and propose a class-aware smoothness optimization algorithm named \underline{Imb}alanced-\underline{SAM} (ImbSAM).
   With the guidance of class priors, our ImbSAM specifically improves generalization targeting tail classes.
   We also verify the efficacy of ImbSAM on two prototypical applications of class-imbalanced recognition: long-tailed classification and semi-supervised anomaly detection, where our ImbSAM demonstrates remarkable performance improvements for tail classes and anomaly. Our code implementation is available at \url{https://github.com/cool-xuan/Imbalanced_SAM}.
\end{abstract}

\begin{figure}[t]
   \centering
   \includegraphics[width=\linewidth]{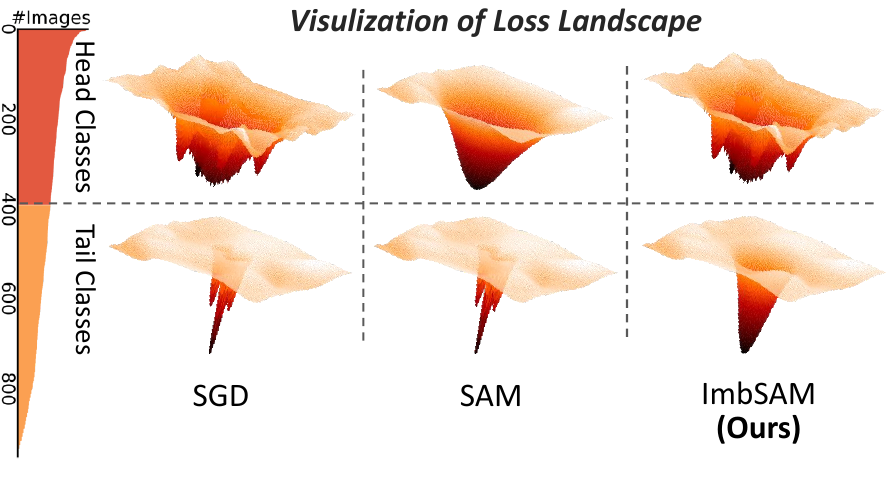}
   \caption{The visualization of separate loss landscape for \emph{head} and \emph{tail} classes in class-imbalanced recognition, optimized by SGD \cite{bottou2012stochastic}, SAM \cite{foret2020sharpness} and our ImbSAM respectively.}
   \label{fig:intro-loss-landscape}
\end{figure}

\section{Introduction}
Over the course of decades, deep neural networks have achieved remarkable success in various already-intensely-studied tasks, including classification \cite{dosovitskiy2020vit, he2016resnet}, segmentation \cite{long2015fcn, ronneberger2015unet}, and object detection \cite{ren2015faster, tian2019fcos}.
These impressive achievements are largely attributed to large-scale datasets \cite{deng2009imagenet, lin2014mscoco}, which strive for a uniform distribution of categories, contrary to the data distribution in the real open world.
The real-world data is usually class-imbalanced \cite{buda2018systematic, he2009learning, krizhevsky2009cifar, liu2019large, van2018inaturalist}, following a long-tailed distribution: a small number of dominant classes (head classes) have numerous samples, while the majority of classes (tail classes) contain only a few samples.
Directly applying SOTA methods \cite{dosovitskiy2020vit, he2016resnet} built under balanced data distribution to the class-imbalanced setting suffers from dramatic performance degradation \cite{liu2019large}.
This critical performance reduction is primarily caused by the overwhelming presence of head classes during training, which in turn results in inadequate learning for tail classes \cite{he2009learning, japkowicz2002class, yang2020rethinking}.
This limitation motivates the theoretical research on class-imbalanced recognition, which drives lots of practical applications such as long-tailed classification \cite{lt2022ltr, kang2019tauNorm, liu2019large, van2018inaturalist} and semi-supervised anomaly detection \cite{han2022adbench, pang2019devnet, ruff2019deep}.

Many excellent methods \cite{cao2019learning, cui2021parametric, cui2019cbloss, drummond2003c4, han2005borderline, lin2017focal, yang2020rethinking} have been proposed to tackle the issue of class-imbalanced data, where plenty of methods re-balance the long-tailed data by re-sampling \cite{drummond2003c4, feng2021exploring} or assign large loss weights to the tail classes.
While these methods alleviate the dominant presence of head classes over tail classes, they overexpose the limited tail class samples, increasing the risk of overfitting for these tail classes \cite{hawkins2004problem}.

Recent methods \cite{lt2022ltr, kang2019tauNorm} fine-tune the regularization to penalize the large parameters in turn avoiding overfitting.
Compared with the empirical regularization, Sharpness-Aware Minimization (SAM) \cite{foret2020sharpness}, an effective optimization algorithm, is supported by a solid theoretical foundation. SAM connects the smooth geometry of the loss landscape with generalization and captures the sharpness of the loss landscape. By simultaneously minimizing the loss value and sharpness, SAM converges the model weights to reach a smooth minimum (neighborhoods having uniformly low loss).

However, the SAM is proposed and effective in the ideal data setting (balanced distribution \cite{deng2009imagenet,krizhevsky2009cifar}), ignoring the class imbalance in the real world.
As the loss landscape visualization of SAM shown in Figure \ref{fig:intro-loss-landscape}, SAM tends to prioritize generalization on the head classes since the heavily imbalanced data, while overlooking the tail classes in class-imbalanced recognition. Nevertheless, even without SAM, the abundant training data of head classes also prevents them from suffering overfitting.

To address this issue, we first investigate the class-imbalanced recognition and identify its generalization bottleneck primarily lying in tail classes.
As for such specific tasks with long-tailed distribution, the head classes, benefiting from sufficient training samples, are less affected by generalization problems \cite{hawkins2004problem}.
On the other hand, the tail classes, with only a few data instances (sometimes even less than 10), are highly susceptible to severe overfitting.
Based on these insights, we propose a class-aware smoothness optimization algorithm named \underline{Imb}alanced-\underline{SAM} (\emph{ImbSAM}) to tackle the overfitting problem with respect to (w.r.t.) tail classes.
In contrast to the class-agnostic SAM, our ImbSAM introduces class priors to restrict the smoothness optimization scope to the tail classes as illustrated in Figure \ref{fig:intro-loss-landscape}, thereby alleviating severe overfitting of inadequate training samples of tail classes.

\begin{figure}[t]
   \centering
   \includegraphics[width=\linewidth]{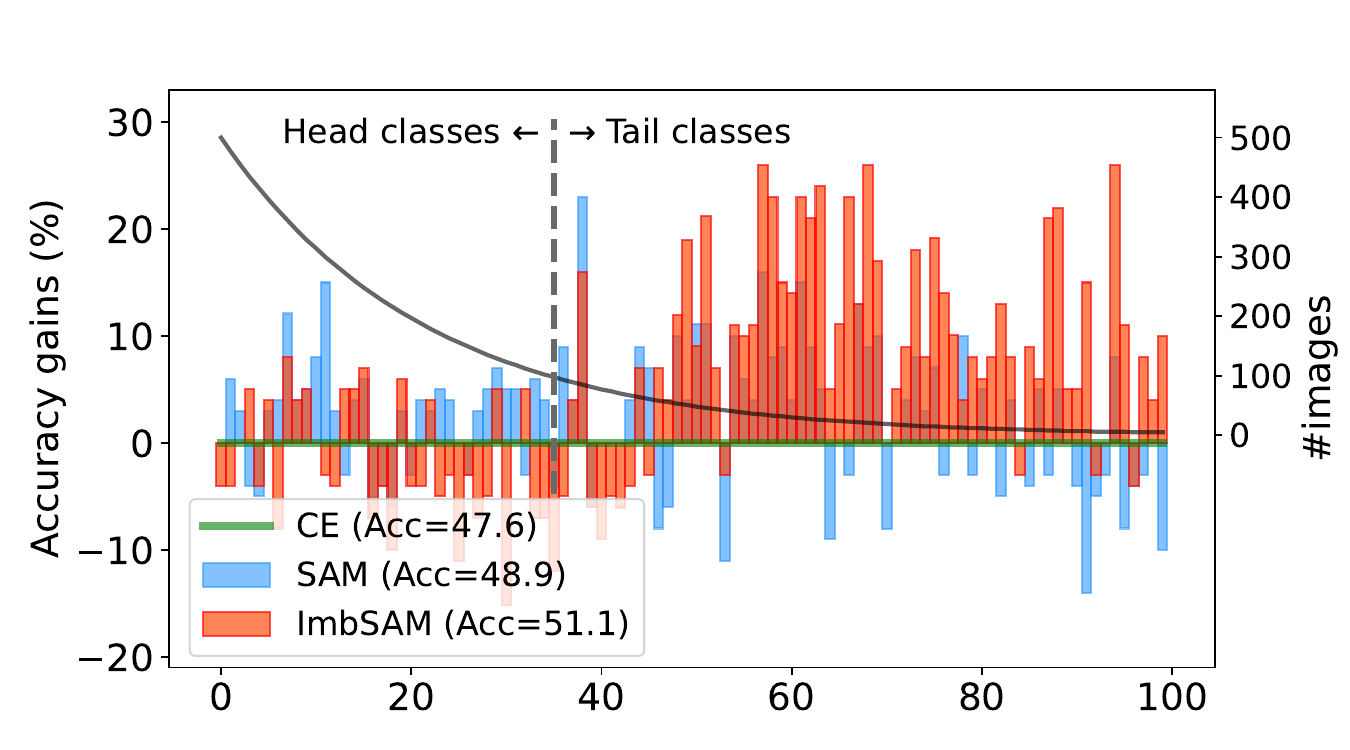}
   \caption{\tbf{Accuracy gains ($\%$) of each classes on CIFAR100-LT} derived from the standard SAM \cite{foret2020sharpness} and our class-aware ImbSAM. With the guidance of class priors (dividing data into two splits by $\eta$), our ImbSAM successfully performs generalization targeting tail classes, which are neglected by SAM.}
   \label{fig:sam-vs-adsam-cifar100}
\end{figure}

Our ImbSAM is compatible with existing methods, demonstrating remarkable performance promotion in prototypical applications of class-imbalanced recognition: long-tailed classification (LTC) \cite{lt2022ltr, he2016resnet, kang2019tauNorm} and semi-supervised anomaly detection (SSAD) \cite{han2022adbench, pang2019devnet, ruff2019deep}. Notably, our ImbSAM impressively improves recognition accuracy for tail classes as illustrated in Figure \ref{fig:sam-vs-adsam-cifar100}, which firmly verifies the efficacy of ImbSAM in focusing generalization scope on classes with limited training data.
Our main contributions are summarized as follows:
\begin{itemize}
   \item We approach the challenge of class-imbalanced recognition from a generalization perspective and identify severe overfitting on tail classes as the main generalization bottleneck. A theoretical analysis is provided to reveal why the promising SAM fails to address the generalization issues in class-imbalanced recognition.
   \item To overcome the limitation of SAM, we propose the Imbalanced SAM (\emph{ImbSAM}), which incorporates class priors into the class-agnostic SAM to specifically address the overfitting problem on tail classes.
   \item We evaluate the efficacy of ImbSAM on two prototypical applications of class-imbalanced recognition: long-tailed classification and semi-supervised anomaly detection, where it demonstrates remarkable performance improvements for the classes with inadequate training samples.
\end{itemize}

\begin{figure*}[t]
   \centering
   \subfigure[SGD]{\includegraphics[width=0.3\textwidth]{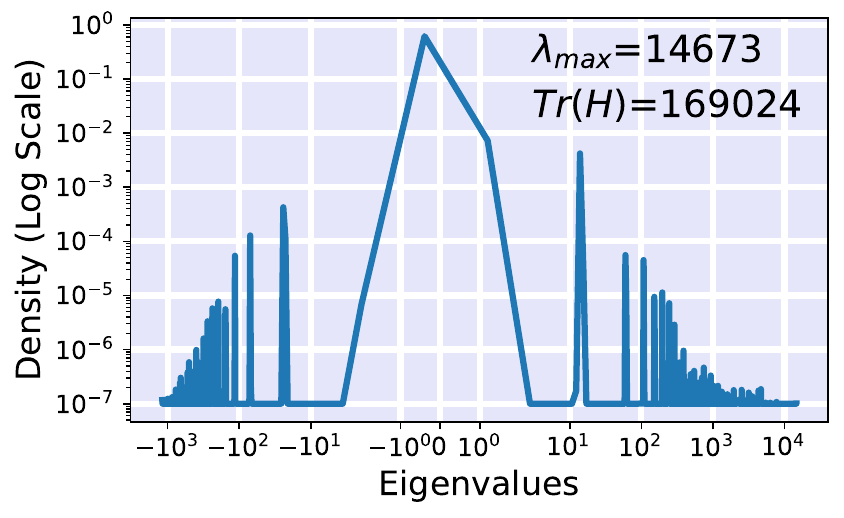}}
   \hspace{0.03\textwidth}
   \subfigure[SAM]{\includegraphics[width=0.3\textwidth]{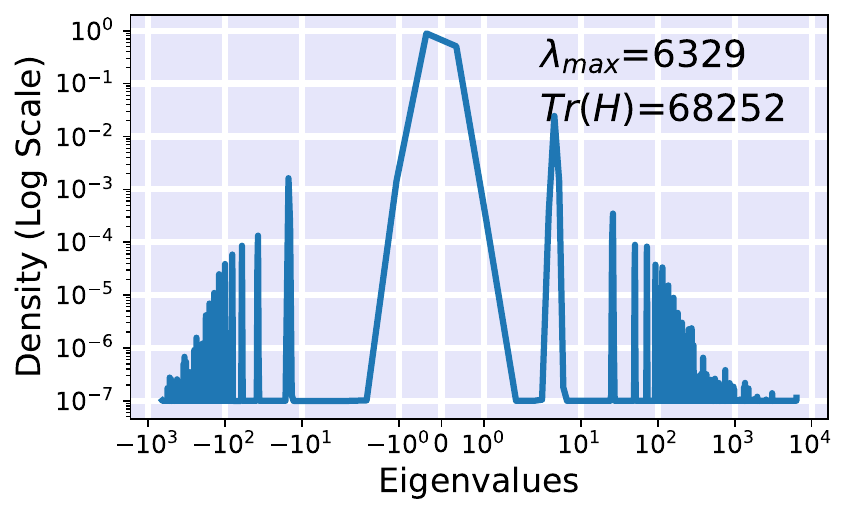}}
   \hspace{0.03\textwidth}
   \subfigure[Our ImbSAM]{\includegraphics[width=0.3\textwidth]{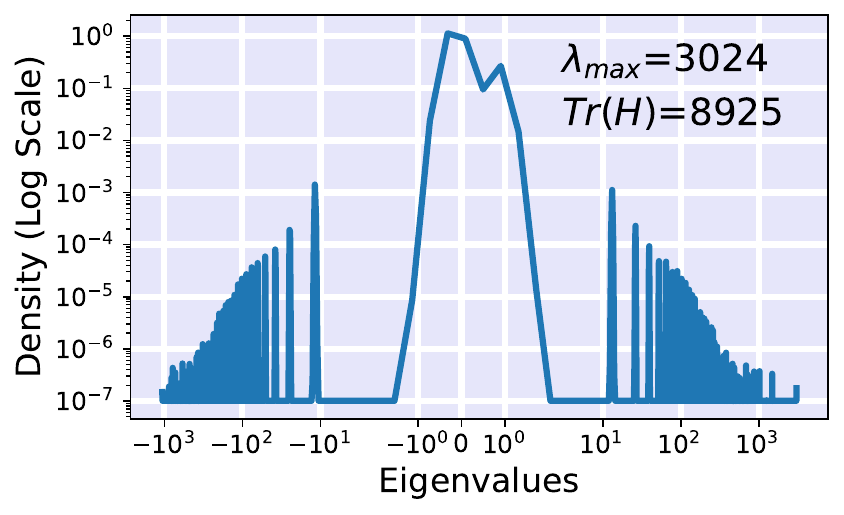}}
   \caption{\tbf{Eigen Spectral Density of Hessian} for the CE \cite{lt2022ltr} respectively optimized by SGD \cite{bottou2012stochastic}, SAM \cite{foret2020sharpness} and our ImbSAM on \emph{Few} classes in CIFAR100-LT (IF=100). Each graph is annotated with the maximum eigenvalue ($\lambda_{max}$) and the trace of the Hessian matrix ($Tr(H)$), which are indicators of the smoothness of loss landscape (Lower $\lambda_{max}$ and $Tr(H)$ reveal the smoother loss landscape).}
   \label{fig:hessian}
\end{figure*}

\section{Related Work}
In this section, two prototypical applications derived from class-imbalanced recognition are first introduced: long-tailed classification and semi-supervised anomaly detection, followed by the introduction of smoothness of loss landscape and its effective implementation: sharpness-aware minimization (SAM).

\ssubsec{Long-Tailed Classification.}
Long-tail classification has been extensively studied \cite{buda2018systematic, haixiang2017learning, he2009learning, japkowicz2002class, johnson2019survey} in recent years due to its importance in real-world applications with heavily imbalanced data distribution.
Early approaches for long-tail classification focused on
re-sampling \cite{drummond2003c4, feng2021exploring} or
re-weighting \cite{cui2019cbloss, lin2017focal} strategies to balance the class distribution,
such as over-sampling tail classes \cite{han2005borderline}
and under-sampling head classes \cite{liu2008exploratory},
or using weighted cross-entropy loss to focus on underrepresented classes \cite{cao2019learning, samuel2021distributional}.
Recently, more flexible and robust methods are proposed involving
transfer learning \cite{liu2018open, yin2019feature},
self-supervision \cite{li2021self},
contrastive learning \cite{cui2021parametric},
representation learning \cite{kang2019tauNorm},
and ensemble learning with multi-experts separately recognize relatively balanced sub-groups \cite{cai2021ace, wang2020long}.

Some existing methods \cite{lt2022ltr,wang2022balanced} also improve long-tailed classification with the perspective of generalization and address the generalization problem by tuning regularization.
However, these methods are more empirical and less controllable compared with Sharpness-Aware Minimization \cite{foret2020sharpness}, which is supported by a solid theoretical foundation.

\ssubsec{Semi-Supervised Anomaly Detection.}
Anomaly detection plays a critical role in broad applications such as violence detection \cite{Purwanto_2021_ICCV} and industrial manufacture \cite{Zavrtanik_2021_ICCV}.
In recent years, semi-supervised anomaly detection methods have shown great potential by leveraging both limited anomaly annotation \cite{akcay2019ganomaly, pang2018learning, pang2019deep, ruff2019deep, Sheynin_2021_ICCV, zhou2021feature} and statistic information of abundant normal data, outperforming traditional unsupervised anomaly detection methods \cite{han2022adbench}.

On the other hand, semi-supervised anomaly detection can be viewed as a special case of long-tailed classification, with only one head class (normal) and one tail class (anomaly), which poses unique challenges since the unpredictable and diverse types of anomalies.
To address this problem, existing semi-supervised anomaly detection methods also borrow the re-sampling strategy \cite{pang2019devnet} or ensemble learning \cite{zhao2018xgbod} from long-tailed classification to balance the extremely long-tailed data. However, the generalization for limited but diverse anomalies remains an open and crucial problem for semi-supervised anomaly detection methods to achieve high recognition accuracy and detect unseen anomalies.

\ssubsec{Smoothness of Loss Landscape.}
The issue of model generalization in deep learning has always been an essential yet challenging problem \cite{caruana2000overfitting, neyshabur2017exploring}. A panoply of outstanding methods have been developed from the scope of model adjustment \cite{ba2016layer, ioffe2015batch} or data augmentation \cite{jiang2019fantastic, zhang2017mixup}.
In recent years, with the perspective of the connection between the geometry of the loss landscape and generalization, the Sharpness-Aware Minimization (SAM) \cite{foret2020sharpness} has emerged as a promising generalization improvement approach by identifying and minimizing the sharpness of the loss landscape.

SAM and its variants \cite{kwon2021asam, rangwani2022sdat} have achieved SOTA performance on various challenging tasks \cite{deng2009imagenet, peng2019moment}, while its application in class-imbalanced recognition remains nascent \cite{rangwani2022escaping, wang2022balanced}. To tackle the generalization issues in class-imbalanced recognition, we enhance SAM with class awareness by incorporating class priors \cite{hand2001simple} and introduce the class-aware ImbSAM. Our ImbSAM successfully narrows down the generalization scope of the standard SAM, specifically targeting the tail classes, which suffer from intractable overfitting.

\section{Proposed Methodology}

\subsection{Notation and Problem Definition}

In class-imbalanced recognition tasks, the training set $\mcal{S}=\cup_{i=1}^n\{(\vx_i, y_i)\}$ is heavily imbalanced, where $y_i \in [1,...,K]$ is labeled class $k$ for the data sample $\vx_i$.
According to the data amount of class $k$, the whole set $\mcal{S}$ is divided into two parts: $\mcal{S}^\trm{head}$ including data of head classes and $\mcal{S}^\trm{tail}$ including data of tail classes, and $\mcal{S}=\mcal{S}^\trm{tail} \cup \mcal{S}^\trm{head}$.
For convenience, $\mcal{S}^k$ denotes the set of training samples belonging to the class $k$, and $|\mcal{S}^k|$ refers to its data amount.
To quantitatively measure how imbalanced the long-tailed dataset $\mcal{S}$ is, the imbalanced factor $\trm{IF}=\frac{\trm{max}_k|\mcal{S}^k|}{\trm{min}_k|\mcal{S}^k|}$ is defined, with $\trm{IF}\gg1$ in class-imbalanced training set $\mcal{S}$.

Class-imbalanced recognition intrinsically follows the common classification framework: training a neural network $f(\cdot\,; \sbf{\theta})$ parameterized by $\sbf{\theta}$.
Given a sample $\vx_i$, the neural network $f$ predicts a label $y^\prime_i\trm{=}f(\vx_i; \sbf{\theta})$.
The classic cross-entropy (CE) loss function \cite{bishop2006pattern} or its variants \cite{cui2019cbloss, lin2017focal} is chosen as the criterion $\ell(y^\prime_i, y_i)$ to supervise the optimization of parameters $\sbf{\theta}$. The training optimization can be formulated as follows:
\begin{align}
   \sbf{\theta^*}
    & = \argmin_{\sbf{\theta}} \sum_{\mcal{S}} \ell(f(\vx_i; \sbf{\theta}), y_i), \nonumber \\
    & = \argmin_{\sbf{\theta}} \sum_k \sum_{\mcal{S}^k} \ell(f(\vx_i; \sbf{\theta}), y_i), \\
    & = \argmin_{\sbf{\theta}} \sum_{\mcal{S}^\trm{head}} \ell(f(\vx_i; \sbf{\theta}), y_i) + \sum_{\mcal{S}^\trm{tail}} \ell(f(\vx_i; \sbf{\theta}), y_i), \nonumber
\end{align}
where $\sbf{\theta^*}$ is the theoretical optimum of $\sbf{\theta}$, and we also provide a class-wise equivalence. Furthermore, we simplify the class-wise equivalence by dividing all $K$ classes into $2$ types: \emph{head} classes with ample samples and \emph{tail} classes with restricted samples.
These two parts of losses are respectively denoted as follows:
\begin{align}
   \mcal{L}_{\mcal{S}^\trm{head}} (\sbf{\theta})
    & = \sum_{\mcal{S}^\trm{head}} \ell(f(\vx_i; \sbf{\theta}), y_i), \label{equ:4} \\
   \mcal{L}_{\mcal{S}^\trm{tail}} (\sbf{\theta})
    & = \sum_{\mcal{S}^\trm{tail}} \ell(f(\vx_i; \sbf{\theta}), y_i). \label{equ:5}
\end{align}
The summarized loss $\mcal{L}_{\mcal{S}} (\sbf{\theta})$ is rewritten as
\begin{equation} \label{equ:6}
   \mcal{L}_{\mcal{S}} (\sbf{\theta})
   = \sum_{\mcal{S}} \ell(f(\vx_i; \sbf{\theta}), y_i)
   = \mcal{L}_{\mcal{S}^\trm{head}} (\sbf{\theta})
   + \mcal{L}_{\mcal{S}^\trm{tail}} (\sbf{\theta}).
\end{equation}

\subsection{Preliminaries: Sharpness-Aware Minimization} \label{sec:sam}

From the perspective of the connection of smooth loss landscape and generalization, Sharpness-Aware Minimization (SAM) not only minimizes the single point in the loss landscape of criterion $\mcal{L}_\mcal{D}(\sbf{\theta})$ w.r.t. data distribution $\mcal{D}$ but also consistently brings its neighborhoods down. As a result, SAM turns to minimize the following PAC-Bayesian error upper bound:
\begin{equation} \label{equ:7}
   \begin{aligned}
      \mcal{L}_\mcal{D}(\sbf{\theta})\leq
       & \left[\max_{\left\lVert \sbf{\epsilon} \right\rVert \leq \rho} \mcal{L}_\mcal{S}(\sbf{\theta} + \sbf{\epsilon}) - \mcal{L}_\mcal{S}(\sbf{\theta})\right]
      \\
       & + \mcal{L}_\mcal{S}(\sbf{\theta})
      + h ({\left\lVert \sbf{\theta} \right\rVert}_2^2/\rho^2),
   \end{aligned}
\end{equation}
with some strictly increasing function $h$.
Compared with standard loss, the additional term in square brackets measures the loss sharpness by capturing the loss increasing rate when perturbing $\sbf{\theta}$ with noise $\sbf{\epsilon}$ in the neighborhood of $\rho$.
Since the monotonicity of $h$, it can be theoretically replaced by the L2 regularization term $\lambda {\left\lVert \sbf{\theta} \right\rVert}_2^2$ with weight decay coefficient $\lambda$.
Thus, the optimization target of SAM, the right-hand side of the inequality Eq. \ref{equ:7}, is rewritten as
\begin{equation} \label{equ:8}
   \min_{\sbf{\theta}}
   \max_{\left\lVert \sbf{\epsilon} \right\rVert \leq \rho} \mcal{L}_\mcal{S}(\sbf{\theta} + \sbf{\epsilon})
   + \lambda {\left\lVert \sbf{\theta} \right\rVert}_2^2,
\end{equation}
which is transferred as a minimax optimization.

To solve this minimax problem, SAM first tackles the max problem by seeking the maximum perturbation $\sbf{\epsilon}_t$ in the range of $\rho$ at training step $t$. This inner maximization problem can be calculated via a first-order Taylor approximation w.r.t. $\sbf{\epsilon}\rightarrow 0$ and dual norm as follows:
\begin{equation} \label{equ:9}
   \begin{aligned}
      \sbf{\epsilon}_t
       & =
      \argmax_{\left\lVert \sbf{\epsilon} \right\rVert \leq \rho} \mcal{L}_\mcal{S} (\sbf{\theta}_t + \sbf{\epsilon})              \\
       & \approx
      \argmax_{\left\lVert \sbf{\epsilon} \right\rVert \leq \rho} \sbf{\epsilon} ^\top \nabla   \mcal{L}_\mcal{S} (\sbf{\theta}_t) \\
       & =
      \rho \trm{sign}(\nabla \mcal{L}_\mcal{S} (\sbf{\theta}_t)) \frac{{\left\lvert \nabla \mcal{L}_\mcal{S} (\sbf{\theta}_t) \right\rvert}^{q-1}}{{\left\lVert \nabla \mcal{L}_\mcal{S} (\sbf{\theta}_t) \right\rVert}^{q/p}_q},
   \end{aligned}
\end{equation}
where ${\left\lvert \cdot \right\rvert}^{q-1}$ refers to element-wise absolute value and power, sign$(\cdot)$ is the signum function, and $1/p+1/q=1$.
Secondly, the outer minimization problem can be solved as
\begin{equation} \label{equ:10}
   \begin{aligned}
      \sbf{\theta}_t
       & = \argmin_{\sbf{\theta}} \mcal{L}_\mcal{S} (\sbf{\theta} + \sbf{\epsilon}_t)
      + \lambda {\left\lVert \sbf\theta \right\rVert}_2^2
      \\
       & \approx \sbf{\theta}_t - \alpha_t \left[\nabla \mcal{L}_\mcal{S} (\sbf{\theta}_t + \sbf{\epsilon}_t) + \lambda \sbf{\theta}_t \right],
   \end{aligned}
\end{equation}
where $\alpha_t$ is the learning rate at training step $t$.
It is empirically confirmed that the above $2$-step optimization yields the best performance when $p=2$, resulting in $\sbf{\epsilon}_t$ formulated as
\begin{equation} \label{equ:11}
   \sbf{\epsilon}_t = \rho \frac{\nabla \mcal{L}_\mcal{S} (\sbf{\theta}_t)}{{\left\lVert \nabla \mcal{L}_\mcal{S} (\sbf{\theta}_t) \right\rVert}_2}.
\end{equation}
In summary, SAM converges $\sbf{\theta}$ to a smooth minimum with uniformly low loss by iteratively solving Eq. \ref{equ:11} and Eq. \ref{equ:10}.

\subsection{The Limitation of Class-Agnostic SAM} \label{sec:sam-limitation}

While SAM is effective and supported by a solid theoretical foundation, the class-agnostic SAM forfeits its impressive generalization power when confront with class-imbalanced data. In long-tailed datasets, the SAM optimization can be re-formulated by introducing the split loss function (Eq. \ref{equ:6}) as follows:
\begin{equation} \label{equ:12}
   \begin{aligned}
      \min_{\sbf{\theta}}
      \max_{\left\lVert \sbf{\epsilon} \right\rVert \leq \rho}
      \big[
       & \mcal{L}_{\mcal{S}^\trm{head}}(\sbf{\theta} + \sbf{\epsilon})   \\
       & + \mcal{L}_{\mcal{S}^\trm{tail}}(\sbf{\theta} + \sbf{\epsilon})
      \big]
      + \lambda {\left\lVert \sbf{\theta} \right\rVert}_2^2.
   \end{aligned}
\end{equation}
Correspondingly, the formula of perturbation $\sbf{\epsilon}_t$ is rewritten as follows:

\begin{equation}
   \quad\sbf{\epsilon}_t = \sbf{\epsilon}_t^\trm{head} + \sbf{\epsilon}_t^\trm{tail}, 
\end{equation}
and 
\begin{equation}
   \begin{aligned}
      \sbf{\epsilon}_t^\trm{head} =
        \rho \frac{\nabla \mcal{L}_{\mcal{S}^\trm{head}} (\sbf{\theta}_t)}
      {{\left\lVert \nabla \mcal{L}_\mcal{S} (\sbf{\theta}_t) \right\rVert}_2}, \quad
      \sbf{\epsilon}_t^\trm{tail} =
      \rho \frac{\nabla \mcal{L}_{\mcal{S}^\trm{tail}} (\sbf{\theta}_t)}
      {{\left\lVert \nabla \mcal{L}_\mcal{S} (\sbf{\theta}_t) \right\rVert}_2},
   \end{aligned}
\end{equation}
where $\sbf{\epsilon}^\trm{head}$ and $\sbf{\epsilon}^\trm{tail}$ denotes the perturbations added for the head class set $\mcal{S}^\trm{head}$ and tail class set $\mcal{S}^\trm{tail}$, respectively.
Due to the overwhelming data amount of head classes over tail classes ($|\mcal{S}^\trm{head}| \gg |\mcal{S}^\trm{tail}|$), the magnitude of the gradient for head classes $|\nabla \mcal{L}_{\mcal{S}^\trm{head}}|$ also crushes $|\nabla \mcal{L}_{\mcal{S}^\trm{tail}}|$ derived from tail classes, resulting in $|\sbf{\epsilon}_t^\trm{head}| \gg |\sbf{\epsilon}_t^\trm{tail}|$.
On the other hand, ${\left\lVert \sbf{\epsilon}_t^\trm{head} + \sbf{\epsilon}_t^\trm{tail} \right\rVert}_2$ equals to $\rho$, which is a constant during training and set $0.05$ for most cases \cite{foret2020sharpness}.
Therefore, the perturbation $\sbf{\epsilon}_t^\trm{tail}$ calculated for tail classes is negligible and can be ignored, which leads to the following approximation of Eq. \ref{equ:12}:
\begin{equation} \label{equ:13}
   \begin{aligned}
      \min_{\sbf{\theta}}
      \max_{\left\lVert \sbf{\epsilon}^\trm{head} \right\rVert \leq \rho}
      \big[
       & \mcal{L}_{\mcal{S}^\trm{head}}(\sbf{\theta} + \sbf{\epsilon}^\trm{head})   \\
       & + \mcal{L}_{\mcal{S}^\trm{tail}}(\sbf{\theta} + \sbf{\epsilon}^\trm{head})
      \big]
      + \lambda {\left\lVert \sbf{\theta} \right\rVert}_2^2.
   \end{aligned}
\end{equation}
Accordingly, the gradient update formula is also approximated as
\begin{equation}
   \begin{aligned}
      \sbf{\theta}_t
      \approx \sbf{\theta}_t
      - \alpha_t \big[
       & \nabla \mcal{L}_{\mcal{S}^\trm{head}} (\sbf{\theta}_t + \sbf{\epsilon}_t^\trm{head})    \\
       & + \nabla \mcal{L}_{\mcal{S}^\trm{tail}} (\sbf{\theta}_t + \sbf{\epsilon}_t^\trm{head}) +\lambda \sbf{\theta}_t \big],
   \end{aligned}
\end{equation}
where the SAM optimization on head classes is persevered, while that on tail classes is misleading by the overwhelming perturbation $\sbf{\epsilon}_t^\trm{head}$.
In other words, the class-agnostic SAM only prioritizes generalization for the head classes, while confusing the optimization of tail classes, which are prone to overfitting.

Introducing re-sampling \cite{han2005borderline, feng2021exploring} or re-weighting \cite{lin2017focal,cui2019cbloss} methods forces the magnitude of $\sbf{\epsilon}^\trm{tail}$ to be larger and even comparable with $\sbf{\epsilon}^\trm{head}$.
At the same time, these two categories of methods also increase the risk of overfitting with limited tail class instances that are over-exposed or over-focused during training.
Notably, SAM is impressive for its generalization improvement, which is theoretically contradictory with re-sampling and re-weighting. The empirical evaluation (Table \ref{tab:comparison-cifar100}) of naively combining SAM with large loss re-weights for tail classes also validates our analysis, where the recognition accuracy of tail classes is barely unaffected even assigned with large re-weight up to 20.

\begin{algorithm}[h]
   \caption{Our \emph{ImbSAM} Algorithm ($p$=$2$)}\label{algor:imbSAM}
   \KwIn{
      Training dataset $\mcal{S}=\cup_{i=1}^n\{(\vx_i, y_i)\}$,
      neural network $f(\cdot)$ with parameters $\sbf{\theta}$,
      loss function $\ell$,
      mini-batch size $b$,
      learning rate $\alpha$,
      weight decay coefficient $\lambda$,
      neighborhood size $\rho$,
      class split threshold $\eta$.}
   \KwOut{Trained parameters $\sbf{\theta}^*$}
   Initialize parameters $\sbf{\theta}_0$, $t=0$; \\
   \While{\textit{not converged}}
   {
   Sample batch $\mcal{B}=\{(\vx_i, y_1), ..., (\vx_b, y_b)\}$; \\
   Divide $\mcal{B}$ into $\mcal{B}^\trm{head}$ and $\mcal{B}^\trm{tail}$ with $\eta$; \quad// Eq. \ref{equ:class_split} \\
   $\mcal{L}_{\mcal{B}^\trm{head}} (\sbf{\theta_t})
      = \sum_{\mcal{B}^\trm{head}} \ell(f(\vx_i; \sbf{\theta}_t), y_i)$; \\
   $\mcal{L}_{\mcal{B}^\trm{tail}} (\sbf{\theta_t})
      = \sum_{\mcal{B}^\trm{tail}} \ell(f(\vx_i; \sbf{\theta}_t), y_i)$; \\
   $\nabla \mcal{L}_{\mcal{B}^\trm{head}} (\sbf{\theta}_t)=\trm{Backward}(\mcal{L}_{\mcal{B}^\trm{head}}, f(\cdot))$; \\
   $\nabla \mcal{L}_{\mcal{B}^\trm{tail}} (\sbf{\theta}_t)=\trm{Backward}(\mcal{L}_{\mcal{B}^\trm{tail}}, f(\cdot))$; \\
   $\sbf{\epsilon}^\trm{tail} =
      \rho \frac{\nabla \mcal{L}_{\mcal{B}^\trm{tail}} (\sbf{\theta}_t)}
      {{\left\lVert \nabla \mcal{L}_\mcal{\mcal{B}^\trm{tail}} (\sbf{\theta}_t) \right\rVert}_2};$ \\
   \small$\sbf{\theta}_t
      = \sbf{\theta}_t
      - \alpha_t \big[
      \nabla \mcal{L}_{\mcal{B}^\trm{tail}} (\sbf{\theta}_t + \sbf{\epsilon}_t^\trm{tail})
      +  \nabla \mcal{L}_{\mcal{B}^\trm{head}} (\sbf{\theta}_t) +\lambda \sbf{\theta}_t \big]$.
   }
\end{algorithm}

\subsection{Class-Aware Imbalanced SAM (ImbSAM)}
Before making some adjustments in the standard SAM to adapt it to class-imbalanced recognition, we first thoroughly investigate this specific task and try to diagnose its generalization bottleneck.
In the class-imbalanced dataset with a long-tailed distribution, a small number of head classes carve up the main body of training data, leaving a limited number of samples for the vast majority of tail classes.
Benefiting from the abundance of training data, there are no severe generalization issues for the minority of head classes \cite{hawkins2004problem}.
However, since tail classes have access to only a handful of training instances, with some classes having fewer than ten samples, deep neural networks are susceptible to overfitting these few training instances and failing to generalize to unseen data.
This results in a generalization bottleneck that is particularly pronounced in tail classes.

Inspired by the above insights, we introduce the class priors to the standard SAM to take full advantage of its efficiently improving generalization and propose a class-aware \underline{Imb}alanced \underline{SAM} (\emph{ImbSAM}) to address the generalization bottleneck on the side of tail classes in class-imbalanced recognition.
With the guidance of class priors, our ImbSAM successfully shifts the focus from dominant head classes to vulnerable tail classes and significantly avoids overfitting.

\ssubsec{How to Build Class Priors.} Due to the lack of class prior that is essential in recognition of tail classes \cite{cai2021ace, cui2019cbloss, haixiang2017learning, lin2017focal,wang2022balanced}, SAM fails to optimize a smooth minimum for tail classes.
Thus, how to construct the class priors for SAM should be first solved.
In our ImbSAM, the class priors are simply built by introducing a class split threshold $\eta$ to divide the entire training set into two sub-sets according to their training data amount: head sub-set $\mcal{S}^\trm{head}$ and tail sub-set $\mcal{S}^\trm{tail}$.
In particular, head and tail sub-sets comprise all data samples categorized into the class with data amount more or not more than $\eta$ respectively, formulated as
\begin{equation} \label{equ:class_split}
   \begin{cases}
      (\vx_i, y_i) \in \mcal{S}^\trm{head} \,\,\,  & \trm{if} \,\,\, |\mcal{S}^{y_i}| > \eta\,          \\
      (\vx_j, y_j)  \in \mcal{S}^\trm{tail} \,\,\, & \trm{if} \,\,\, |\mcal{S}^{y_j}| \leqslant \eta\,,
   \end{cases}
\end{equation}
where the class split threshold $\eta$ is a hyperparameter to control the training set splitting and can be set $100$ for the most long-tailed datasets \cite{cui2019cbloss,liu2019large,van2018inaturalist}.
Although our class priors only roughly split the entire training set $\mcal{S}$ into two parts without considering the specific data amount of each class, they still play a crucial role in endowing our ImbSAM with class awareness.

\ssubsec{How to Utilize Class Priors in ImbSAM.}
To incorporate class awareness into the standard SAM, we treat the losses derived from the two sub-sets divided according to our class priors, $\mcal{L}_{\mcal{S}^\trm{head}} (\sbf{\theta})$ and $\mcal{L}_{\mcal{S}^\trm{tail}} (\sbf{\theta})$, differently.
Particularly, the optimization target in our ImbSAM is adapted from Eq. \ref{equ:12} to the following formula by ignoring the SAM optimization term for head classes:
\begin{equation}
   \min_{\sbf{\theta}}
   \max_{\left\lVert \sbf{\epsilon} \right\rVert \leq \rho}
   \mcal{L}_{\mcal{S}^\trm{tail}}(\sbf{\theta} + \sbf{\epsilon}^\trm{tail}) + 
   \mcal{L}_{\mcal{S}^\trm{head}}(\sbf{\theta})
   + \lambda {\left\lVert \sbf{\theta} \right\rVert}_2^2.
\end{equation}
To make explicit our sharpness-aware term, the above optimization target can be rewritten as follows:
\begin{align}
   & \qquad\qquad\quad\trm{optimization term for \emph{tail} classes} \nonumber
  \\[-5pt]
   & \min_{\sbf{\theta}}
  \overbrace{
  \left[
  \max_{\left\lVert \sbf{\epsilon} \right\rVert \leq \rho}
  \mcal{L}_{\mcal{S}^\trm{tail}}(\sbf{\theta} + \sbf{\epsilon}^\trm{tail})
  - \mcal{L}_{\mcal{S}^\trm{tail}}(\sbf{\theta})
  \right]
  + \mcal{L}_{\mcal{S}^\trm{tail}}(\sbf{\theta})} \nonumber
  \\
   & \qquad+ \underbrace{\mcal{L}_{\mcal{S}^\trm{head}}(\sbf{\theta})}
  + \lambda {\left\lVert \sbf{\theta} \right\rVert}_2^2,
  \\[-5pt]
   & \trm{optimization term for \emph{head} classes} \nonumber
\end{align}
where the term in square brackets specifically captures the sharpness for the loss derived from tail classes. Unlike the class-agnostic SAM treating all classes equally, our ImbSAM leverage class priors to focus the sharpness-aware minimization on tail classes specifically, and maintain the standard optimization for head classes. 

\begin{figure*}[t]
   \begin{minipage}[c]{\textwidth}
      \begin{minipage}[t]{0.35\textwidth}
         \centering
         \captionof{table}{\tbf{Comparison of overall accuracy ($\%$) on CIFAR100-LT} with IF=[100, 50, 10]. The reported accuracy of CE and WB \cite{lt2022ltr} is implemented by ourselves. `*' refers to the SOTA with bells and whistles. }
         \label{tab:comparison-cifar3}
         \resizebox{0.9\textwidth}{!}{
            \begin{tabular}{l|ccc}
               \myrule
               Imbalance Factor                       & 100        & 50         & 10         \\ \hline
               CB \cite{cui2019cbloss}                & 39.6       & 45.2       & 58.0       \\
               KD \cite{hinton2015distilling}         & 40.4       & 45.5       & 59.2       \\
               LDAM-DRW \cite{cao2019learning}        & 42.0       & 46.6       & 58.7       \\
               BBN \cite{zhou2020bbn}                 & 42.6       & 47.0       & 58.7       \\
               De-confound \cite{tang2020long}        & 44.1       & 50.3       & 59.6       \\
               $\tau$-norm \cite{kang2019tauNorm}     & 47.7       & 52.5       & 63.8       \\
               DiVE \cite{he2021distilling}           & 45.4       & 51.1       & 62.0       \\
               DRO-LT \cite{samuel2021distributional} & 47.3       & 57.6       & 63.4       \\
               SSD* \cite{li2021self}                 & 46.0       & 50.5       & 62.3       \\
               PaCO* \cite{cui2021parametric}         & 52.0       & 56.0       & 64.2       \\
               ACE* \cite{cai2021ace}                 & 49.6       & 51.9       & $-$        \\ \hline
               CE \cite{lt2022ltr}                    & 47.6       & 52.8       & 66.9       \\
               \rowcolor[HTML]{e6e6e6}
               CE+\tbf{ImbSAM}                        & \tbf{51.1} & \tbf{56.4} & \tbf{69.2} \\ \mytinyrule
               WB \cite{lt2022ltr}                    & 51.9       & 56.7       & 68.9       \\
               \rowcolor[HTML]{e6e6e6}
               WB+\tbf{ImbSAM}                        & \tbf{54.8} & \tbf{59.3} & \tbf{69.7} \\ \myrule
            \end{tabular}}
      \end{minipage}
      \hspace{10pt}
      \begin{minipage}[t]{0.64\textwidth}
         \centering
         \captionof{table}{\tbf{Comparison of overall accuracy and split accuracy ($\%$) on large-scale ImageNet-LT and iNaturalist}. The reported accuracy of CE \cite{lt2022ltr} and LWS \cite{kang2019tauNorm} is implemented by ourselves. `*' refers to the SOTA with bells and whistles. The unreported accuracy in \cite{samuel2021distributional} and \cite{li2021self} is replaced with `-'. `Med.' denotes `Medium' classes.}
         \label{tab:comparison-imagenet-inat}
         \resizebox{0.865\textwidth}{!}{
            \begin{tabular}{lcccccccc}
               \myrule
                                                                           & \multicolumn{4}{c}{ImageNet-LT \cite{liu2019large}} & \multicolumn{4}{c}{iNaturaList \cite{van2018inaturalist}}                                                                               \\
               \cmidrule(r){2-5} \cmidrule(r){6-9}
                                                                           & Many                                                & Med.                                                      & Few        & All        & Many       & Med.       & Few        & All        \\ \hline
               \multicolumn{1}{l|}{CB \cite{cui2019cbloss}}                & 39.6                                                & 32.7                                                      & 16.8       & 33.2       & 53.4       & 54.8       & 53.2       & 54.0       \\
               \multicolumn{1}{l|}{$\tau$-norm \cite{kang2019tauNorm}}     & 59.1                                                & 46.9                                                      & 30.7       & 49.4       & 65.6       & 65.3       & 65.5       & 65.6       \\
               \multicolumn{1}{l|}{DiVE \cite{he2021distilling}}           & 64.1                                                & 50.4                                                      & 31.5       & 53.1       & 70.6       & 70.0       & 67.7       & 69.1       \\
               \multicolumn{1}{l|}{DRO-LT \cite{samuel2021distributional}} & 64.0                                                & 49.8                                                      & 33.1       & 53.5       & $-$        & $-$        & $-$        & 69.7       \\
               \multicolumn{1}{l|}{DisAlign \cite{zhang2021distribution}}  & 61.3                                                & 52.2                                                      & 31.4       & 52.9       & 69.0       & 71.1       & 70.2       & 70.6       \\
               \multicolumn{1}{l|}{WB \cite{lt2022ltr}}                    & 62.5                                                & 50.4                                                      & 41.5       & 53.9       & 71.2       & 70.4       & 69.7       & 70.2       \\
               \multicolumn{1}{l|}{PaCO* \cite{cui2021parametric}}         & 63.2                                                & 51.6                                                      & 39.2       & 54.4       & 69.5       & 72.3       & 73.1       & 72.3       \\
               \multicolumn{1}{l|}{SSD* \cite{li2021self}}                 & 66.8                                                & 53.1                                                      & 35.4       & 56.0       & $-$        & $-$        & $-$        & 71.5       \\
               \multicolumn{1}{l|}{RIDE* \cite{wang2020long}}              & 67.9                                                & 52.3                                                      & 36.0       & 56.1       & 66.5       & 72.1       & 71.5       & 71.3       \\ \hline
               \multicolumn{1}{l|}{CE \cite{lt2022ltr}}                    & 69.3                                                & 41.7                                                      & 10.3       & 48.2       & 75.4       & 66.9       & 61.7       & 65.7       \\
               \multicolumn{1}{l|}{CE+SAM \cite{foret2020sharpness}}       & \tbf{70.0}                                          & 41.1                                                      & 10.2       & 48.2       & \tbf{75.6} & 66.7       & 61.8       & 65.7       \\
               \rowcolor[HTML]{e6e6e6}
               \multicolumn{1}{l|}{CE+\tbf{ImbSAM}}                        & 68.5                                                & \tbf{47.5}                                                & \tbf{21.6} & \tbf{52.2} & 73.5       & \tbf{69.2} & \tbf{67.9} & \tbf{69.1} \\ \mytinyrule
               \multicolumn{1}{l|}{LWS \cite{kang2019tauNorm}}             & \tbf{64.1}                                          & 49.1                                                      & 31.2       & 52.5       & \tbf{71.7} & 69.4       & 68.7       & 69.4       \\
               \multicolumn{1}{l|}{LWS+SAM \cite{foret2020sharpness}}      & 64.0                                                & 48.8                                                      & 30.5       & 52.3       & \tbf{71.7} & 69.6       & 68.8       & 69.5       \\
               \rowcolor[HTML]{e6e6e6}
               \multicolumn{1}{l|}{LWS+\tbf{ImbSAM}}                       & 63.2                                                & \tbf{53.7}                                                & \tbf{38.3} & \tbf{55.3} & 68.2       & \tbf{72.5} & \tbf{72.9} & \tbf{71.1} \\ \myrule
            \end{tabular}}
      \end{minipage}

   \end{minipage}
\end{figure*}

The gradient update is also changed accordingly as
\begin{equation}
   \begin{aligned}
      \sbf{\theta}_t
      = \sbf{\theta}_t
      - \alpha_t \big[
        & \nabla \mcal{L}_{\mcal{S}^\trm{tail}} (\sbf{\theta}_t + \sbf{\epsilon}_t^\trm{tail}) \\
      + & \nabla \mcal{L}_{\mcal{S}^\trm{head}} (\sbf{\theta}_t) +\lambda \sbf{\theta}_t \big]
   \end{aligned}
\end{equation}
with $\sbf{\epsilon}_t^\trm{tail}$ calculated as
\begin{equation}
   \sbf{\epsilon}^\trm{tail} =
   \rho \frac{\nabla \mcal{L}_{\mcal{S}^\trm{tail}} (\sbf{\theta}_t)}
   {{\left\lVert \nabla \mcal{L}_\mcal{\mcal{S}^\trm{tail}} (\sbf{\theta}_t) \right\rVert}_2}.
\end{equation}
As demonstrated in the above gradient update, the proposed ImbSAM suspends the sharpness-aware minimization for head classes, which are less prone to overfitting with the support of sufficient training data.

By incorporating class priors into the class-agnostic SAM, our ImbSAM efficiently restricts its uncontrollable generalization scope from all classes, which are typically dominated by head classes with overwhelming data, to the overlooked tail classes that are plagued with overfitting problems.
Algorithm \ref{algor:imbSAM} outlines the full ImbSAM algorithm, with SGD as the base gradient optimizer.
Additionally, the pseudo-code in \texttt{PyTorch} style of our ImbSAM is displayed in the supplementary, which only requires a few changes in the implementation of the standard SAM. We also provide the other implementation with the Huawei MindSpore toolkit at \url{https://github.com/cool-xuan/Imbalanced_SAM}.

\section{Experiments}
Comprehensive empirical experiments are conducted to verify the generalization improvement efficacy of our ImbSAM when confront with class-imbalanced data.
Our experiments encompass two prototypical applications: Long-Tailed Classification (LTC) and Semi-Supervised Anomaly Detection (SSAD), demonstrating the broad applicability of the proposed ImbSAM. 
In all experiments, we evaluate the effectiveness of our ImbSAM by simply replacing the original optimization algorithm (SGD \cite{bottou2012stochastic} with momentum \cite{sutskever2013importance}) used in existing methods with our class-aware ImbSAM, without any other hyperparameter changing.

\subsection{Long-Tailed Classification (LTC)}

\ssubsec{Datasets.} We conduct experiments on three mainstream long-tailed datasets including CIFAR100-LT, ImageNet-LT, and iNaturalist. CIFAR100-LT \cite{cui2019cbloss} and ImageNet-LT \cite{liu2019large} are artificially truncated from the balanced CIFAR100 \cite{krizhevsky2009cifar} and ImageNet \cite{deng2009imagenet} datasets, while iNaturalist \cite{van2018inaturalist} is a large-scale naturally imbalanced dataset comprising $8,142$ species with the number of samples per class ranges from $1,000$ to $2$.
ImageNet-LT also has 1000 classes like the balanced version and the number of samples per class ranges from $1,280$ to $5$ images.
Particularly, there are three sub-versions of CIFAR100-LT by varying the imbalanced factor (IF) in $[100, 50, 10]$.

\ssubsec{Evaluation Protocol.} In long-tailed classification, all classes with long-tailed distribution are treated equally during testing. The overall accuracy is calculated as $\trm{acc}_\trm{All}\trm{=}\frac{1}{K} \sum \trm{acc}_k$ including \emph{All} classes, where $\trm{acc}_k$ is the Top-$1$ recognition accuracy for class $k$. Following \cite{liu2019large}, we also report accuracy on three splits of classes according to the number of training data: \emph{Many} classes ($>$100), \emph{Medium} classes (20$\thicksim$100), and \emph{Few} classes ($<$20).

\ssubsec{Implementation.} Our ImbSAM is combined with existing long-tailed classification methods to demonstrate its efficacy, including the baseline trained by cross-entropy loss (CE) with fine-tuned weight decay \cite{lt2022ltr} and two strong SOTA methods: weight balancing (WB) \cite{lt2022ltr} for CIFAR100-LT and learnable weight scaling (LWS) \cite{kang2019tauNorm} for the other two large-scale datasets.
For a fair comparison to prior methods, we use ResNet32 \cite{cui2019cbloss} on CIFAR100-LT, ResNeXt50 \cite{xie2017aggregated} on ImageNet-LT, and ResNet50 \cite{he2016resnet} on iNaturalist2018.
We set SGD optimizer with momentum $0.9$ as the base optimizer and train all models for 200 epochs, with a batch size of $64$ for CIFAR100-LT and ImageNet-LT, and $512$ for iNaturalist.
For all experiments, if not specified, the hyperparameters $\rho$=$0.05$ and $\eta$ is set as $100$, which equals the upper limit of \emph{Medium} classes.

\ssubsec{Comparison on CIFAR100-LT.}
As reported in Table \ref{tab:comparison-cifar3}, our ImbSAM demonstrates consistently significant accuracy improvement for the naive baseline (CE) or strong SOTA (WB \cite{lt2022ltr}) on the CIFAR100-LT dataset. When combined with the prior SOTA WB, our ImbSAM further achieves a novel SOTA performance with overall accuracy $54.8\%$, $59.3\%$ and $69.7\%$ respectively, on all three sub-versions (IF=[100, 50, 10]).

Furthermore, we report the detailed accuracy of \emph{Many}, \emph{Medium}, and \emph{Few} on the most imbalanced CIFAR100-LT with IF=$100$ in Table \ref{tab:comparison-cifar100}, where we also apply the standard SAM ($\rho=0.05$ \cite{foret2020sharpness}) to these methods and coordinate it with large weights for tail classes.
As we claimed in Section \ref{sec:sam-limitation}, the class-agnostic SAM only performs its powerful generalization improvement on the dominant head classes while neglecting the severe overfitting lying in tail classes. Assigning SAM with large loss weights up to $20$ does not yield an obvious corrective effect.
However, with the class priors, our ImbSAM efficiently addresses the overfitting issues for tail classes, resulting in a significant accuracy improvement ($>$4\%) for the \emph{Medium} and \emph{Few} sub-sets without heavily sacrificing the performance on \emph{Many} classes.

\ssubsec{Detailed performance improvements for each class.}
We also visualize the accuracy gains of SAM and our ImbSAM for each class in Figure \ref{fig:sam-vs-adsam-cifar100}, which intuitively displays our ImbSAM's efficacy in avoiding overfitting for the tail classes with barely a handful of training samples.
Besides, we calculate the Eigen Spectral Density \cite{yao2020pyhessian} for CE respectively optimized by SGD \cite{bottou2012stochastic}, SAM and the proposed ImbSAM on \emph{Few} classes in CIFAR100-LT, as illustrated in Figure \ref{fig:hessian}.
Particularly, the maximum eigenvalue ($\lambda_{max}$) and the trace of the Hessian matrix ($Tr(H)$) derived from the model optimized by ImbSAM are the smallest among the model trained by three optimizers, empirically demonstrating the efficacy of our ImbSAM in performing smoothness optimization targeting tail classes.

\begin{table}[t]
   \centering
   \caption{\tbf{Split accuracy ($\%$) comparison on CIFAR100-LT (IF=100)} between SAM \cite{foret2020sharpness} and our ImbSAM. In particular, we assign the class-agnostic SAM with large weights (2, 3, 5, 10, and 20) for tail classes while no re-weighting for our ImbSAM. `Med.' denotes `Medium'.}
   \label{tab:comparison-cifar100}
   \resizebox{0.9\columnwidth}{!}{%
      \begin{tabular}{l|llll}
         \myrule
         Models                            & Many       & Med.       & Few        & All        \\ \hline
         CE \cite{lt2022ltr}               & 77.8       & 46.6       & 13.5       & 47.6       \\
         CE+SAM \cite{foret2020sharpness}  & 79.7       & 49.3       & 12.4       & 48.9       \\
         CE+SAM (reweight=2)               & 79.4       & 49.1       & 12.3       & 48.7       \\
         CE+SAM (reweight=3)               & 79.9       & 48.9       & 13.4       & 49.1       \\
         CE+SAM (reweight=5)               & 79.5       & 49.2       & 13.2       & 49.0       \\
         CE+SAM (reweight=10)              & 79.7       & 48.6       & 13.1       & 48.8       \\
         CE+SAM (reweight=20)              & \tbf{80.1} & 48.7       & 13.5       & 49.1       \\
         \rowcolor[HTML]{e6e6e6}
         CE+\tbf{ImbSAM}                   & 75.9       & \tbf{53.5} & \tbf{19.4} & \tbf{51.1} \\ \hline
         LTR \cite{lt2022ltr}              & 68.5       & 49.1       & 35.9       & 51.9       \\
         LTR+SAM \cite{foret2020sharpness} & \tbf{76.1} & 54.6       & 25.3       & 53.3       \\
         \rowcolor[HTML]{e6e6e6}
         LTR+\tbf{ImbSAM}                  & 64.1       & \tbf{58.6} & \tbf{39.4} & \tbf{54.8} \\ \myrule
      \end{tabular}%
   }
\end{table}

\ssubsec{Comparison on ImageNet-LT and iNaturalist.}
On these two large-scale datasets following irregular and complex data distribution, our ImbSAM also exhibits superior accuracy gains.
Specifically, Table \ref{tab:comparison-imagenet-inat} shows that when combined with the selected baseline (CE \cite{lt2022ltr}) and SOTA (LWS \cite{kang2019tauNorm}), the proposed ImbSAM yields significant accuracy improvements of $3\%\thicksim11\%$ on the \emph{Medium} and \emph{Few} class splits. Although our ImbSAM slightly affects the performance on \emph{Few} classes, it still leads to overall accuracy increases of $1.7\%\thicksim4\%$.
Without bells and whistles, our ImbSAM promotes the prior method LWS \cite{kang2019tauNorm} to be comparable with the prior SOTAs that commonly employ ensemble learning (RIDE \cite{wang2020long}), self-pretraining (PaCO \cite{cui2021parametric} or SSD \cite{li2021self}), and achieves a novel SOTA performance of the trade-off on \emph{Medium} and \emph{Few} classes.

In particular, on these two large-scale datasets containing much more classes ($1,000$ classes for ImageNet-LT and $8,142$ classes for iNaturalist) than the artificial CIFAR100-LT, the standard SAM loses its impressive efficacy in generalization improving even for \emph{Many} classes. This is because the head classes in these datasets have sufficient training data, which inherently guarantees their generalization \cite{hawkins2004problem}. In this case, the classification performance is mainly limited by the model capacity \cite{bommasani2021opportunities}.

\begin{figure}[htbp]
   \setcounter{figure}{4}
   \centering
   \begin{minipage}[t]{0.47\linewidth}
      \centering
      \includegraphics[width=\linewidth]{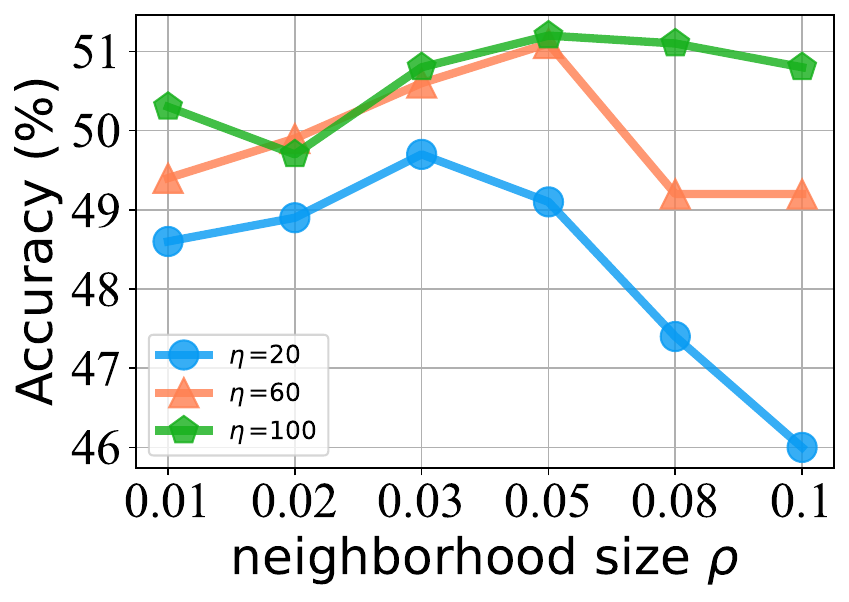}
      \captionof{subfigure}{Overall accuracy.}
      \label{fig:abla-rho-eta:all}
   \end{minipage}
   \begin{minipage}[t]{0.47\linewidth}
      \centering
      \includegraphics[width=\linewidth]{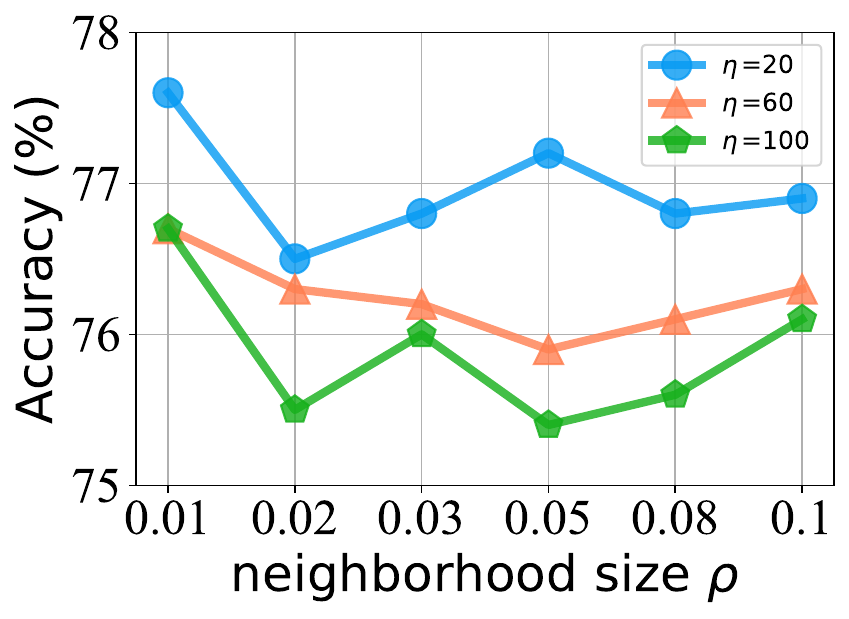}
      \captionof{subfigure}{Accuracy on \emph{Many}.}
      \label{fig:abla-rho-eta:many}
   \end{minipage}
   \begin{minipage}[t]{0.47\linewidth}
      \centering
      \includegraphics[width=\linewidth]{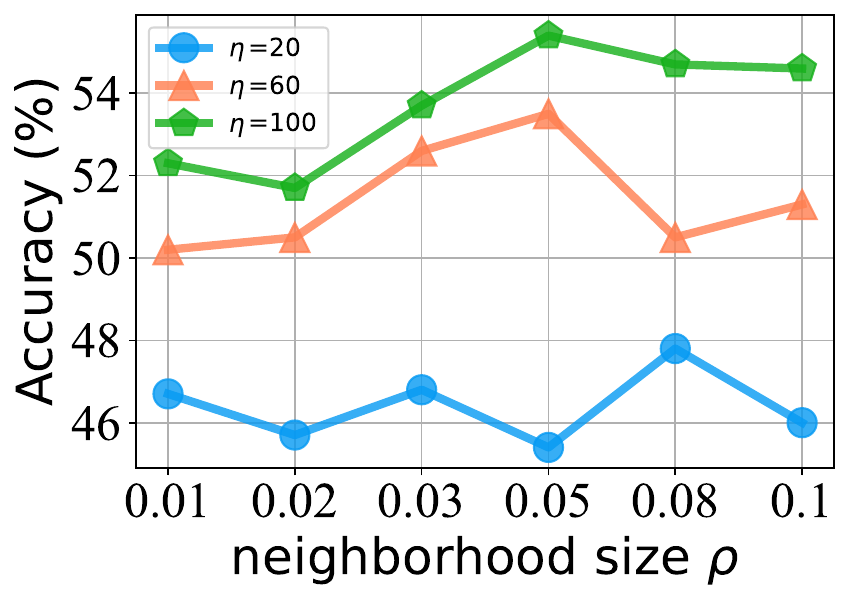}
      \captionof{subfigure}{Accuracy on \emph{Medium}.}
      \label{fig:abla-rho-eta:med}
   \end{minipage}
   \begin{minipage}[t]{0.47\linewidth}
      \centering
      \includegraphics[width=\linewidth]{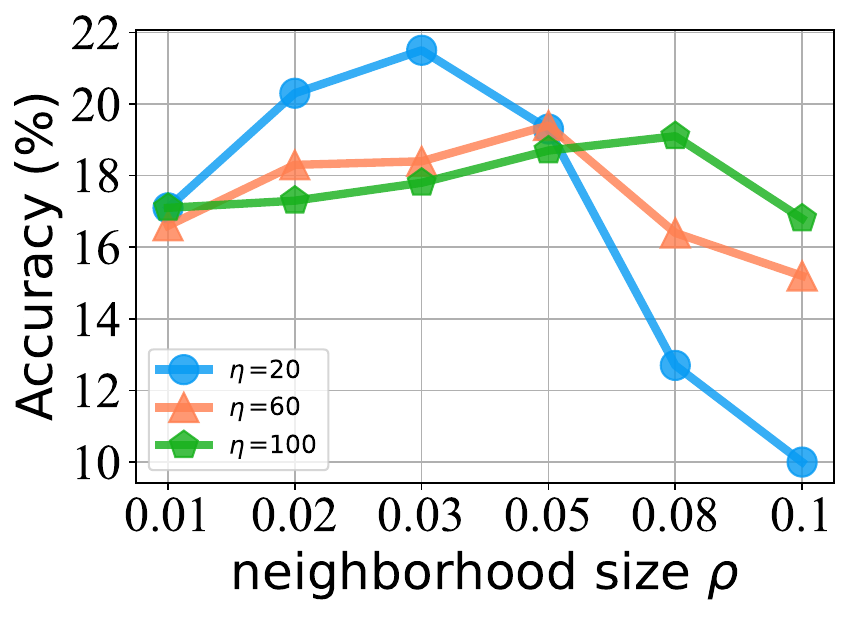}
      \captionof{subfigure}{Accuracy on \emph{Few}.}
      \label{fig:abla-rho-eta:few}
   \end{minipage}
   \setcounter{figure}{3}
   \caption{\tbf{Detailed accuracy ($\%$) variation on CIFAR100-LT (IF=100)} by varying the neighborhood size $\rho$ and class split threshold $\eta$ in our ImbSAM, based on the baseline naive trained with cross-entropy loss (CE) \cite{lt2022ltr}.}
\end{figure}

\ssubsec{Ablation study about $\rho$ and $\eta$ in ImbSAM.}
We conduct the ablation studies by varying neighborhood size $\rho \in$ [0.01, 0.02, 0.03, 0.05, 0.08, 0.1] and $\eta \in$ [20, 60, 100] on CIFAR100-LT with IF=100, based on the naive baseline (CE) optimized with our ImbSAM. According to the overall accuracy variation tendency as shown in Figure \ref{fig:abla-rho-eta:all}, the performance achieves the best with  $\rho$=$0.05$ and class split threshold $\eta$=$100$, which exactly equals to the upper bound of \emph{Medium} classes.
Particularly, for \emph{Few} classes, our ImbSAM boosts CE to achieve a superior accuracy up to $21.5\%$ when $\eta$ is set as the threshold of \emph{Few} classes, which further verifies the controllable generalization scope of our ImbSAM.

Similar to SAM \cite{foret2020sharpness}, $\rho$=$0.05$ is the best hyperparameter setting on \emph{Medium} and \emph{Few} classes, when $\eta$ is set 60 or 100 (Figure \ref{fig:abla-rho-eta:many} and Figure \ref{fig:abla-rho-eta:med}).
However, the best setting for $\rho$ is $0.03$ under $\eta$=$20$ as shown in Figure \ref{fig:abla-rho-eta:few}, and the accuracy on \emph{Few} classes is dramatically reduced with $\rho$ increasing. Since the harsh data limitation ($<20$), the gradient supervision derived from neighborhoods of the large range is noisy, which contributes little to model training.

\begin{table}[h]
   \centering
   \caption{\tbf{Comparison of AUCROC score ($\%$) on five AD datasets} with anomaly ratio $\gamma_l$=25\%. `*' refers to the SOTA with ensembling.}
   \label{tab:comparison-ssad}
   \resizebox{\columnwidth}{!}{%
      \begin{tabular}{l|cccccc}
         \myrule
         Model                              & CIFAR      & F-MNIST    & MNIST-C    & MVTec      & SVHN       & Avg        \\ \hline
         GANomaly \cite{akcay2019ganomaly}  & 67.8       & 79.4       & 75.1       & 76.0       & 56.9       & 71.0       \\
         REPEN  \cite{pang2018learning}     & 67.9       & 87.1       & 80.9       & 75.9       & 58.8       & 74.1       \\
         PReNet \cite{pang2019deep}         & 87.5       & 96.1       & 94.4       & 90.2       & 78.7       & 89.4       \\
         FEAWAD \cite{zhou2021feature}      & 85.2       & 95.1       & 95.4       & 96.2       & 77.4       & 89.9       \\
         XGBOD* \cite{zhao2018xgbod}        & 87.8       & 96.4       & 95.8       & 99.1       & 81.2       & 92.1       \\
         \hline
         DeepSAD \cite{ruff2019deep}        & 86.5       & 96.3       & \tbf{96.4} & 93.1       & 80.9       & 90.6       \\
         \,\,+SAM \cite{foret2020sharpness} & 86.9       & 96.7       & 96.1       & 91.3       & 81.0       & 90.4       \\
         \rowcolor[HTML]{e6e6e6}
         \,\,+\tbf{ImbSAM}                  & \tbf{87.9} & \tbf{97.0} & 96.2       & \tbf{95.3} & \tbf{82.4} & \tbf{91.8} \\ \mytinyrule
         DevNet \cite{pang2019devnet}       & 88.4       & 96.4       & 95.6       & 95.6       & 82.1       & 91.6       \\
         \,\,+SAM \cite{foret2020sharpness} & 88.2       & 96.2       & 95.3       & 94.2       & 81.0       & 91.0       \\
         \rowcolor[HTML]{e6e6e6}
         \,\,+\tbf{ImbSAM}                  & \tbf{88.7} & \tbf{96.7} & \tbf{96.0} & \tbf{96.2} & \tbf{83.7} & \tbf{92.2} \\ \myrule
      \end{tabular}%
   }
\end{table}

\subsection{Semi-Supervised Anomaly Detection (SSAD)}
Compared with long-tailed classification, although there are only one head class (normal) and one tail class (anomaly) in SSAD, the diversity and uncertainty of anomalies make this task challenging.

\ssubsec{Datasets.} Following the impressive ADBench \cite{han2022adbench}, we evaluate our ImbSAM on five image datasets: CIFAR10, SVHN, FashionMNIST, MNIST-C, and MVTec-AD.
The former $4$ datasets respectively contain $10$ sub-sets with one class as normal and other classes as abnormal. For MNIST-C, original MNIST samples are set as normal and corrupted images as abnormal. In MVTec-AD, $15$ types of industrial products are collected with accepts as normal and defects as abnormal.
The number of accessible anomalies during training is controlled by the anomaly ratio $\gamma_l$ following \cite{han2022adbench}.

\ssubsec{Evaluation protocol.} We calculate the widely-used AUCROC (Area Under Receiver Operating Characteristic Curve) and AUCPR (Area Under Precision-Recall Curve) to evaluate the detection accuracy following \cite{han2022adbench}. In particular, we report the AUCPR w.r.t. both normal and anomaly, which can be viewed as the detection performance for normal and anomaly, respectively. Notably, we only report the dataset-wise performance and the average performance of five datasets in the text, detailed performance in each dataset is displayed in our supplementary.

\ssubsec{Implementation.} The SOTA methods DeepSAD \cite{ruff2019deep} and DevNet \cite{pang2019devnet} are selected as two strong baselines to combine with our ImbSAM. For all SSAD datasets, the frozen ResNet18 \cite{he2016resnet} is adopted to extract features. All models are trained for $50$ epochs with a batch size of $128$, and $\rho$ in SAM and ImbSAM is set as $0.05$ for all experiments.

\ssubsec{Comparison on SSAD datasets.}
When applied to SSAD, our ImbSAM simply treats the anomalies as the only tail class and adopts the generalization capacity of SAM targeting the optimization for anomaly perception.
As shown in Table \ref{tab:comparison-ssad}, ImbSAM further boosts the selected two strong baselines, DeepSAD \cite{ruff2019deep} and DevNet \cite{pang2019devnet}, to achieve higher AUCROC scores, outperforming the prior ensembling SOTA (92.1\% of XGBOD \cite{zhao2018xgbod} v.s. 92.2\% of DevNet optimized with our ImbSAM). In particular, DevNet incorporated with our ImbSAM achieves superior performance over XGBOD on all datasets except for MVTec AD, an industrial defect detection dataset. As for this specific dataset, XGBOD benefits from the prediction of ensemble unsupervised methods only trained by normal samples, thoroughly capturing the characteristics of defect-free data, leading to the best performance.

\begin{figure}[t]
   \centering
   \subfigure[\label{fig:abla-ssad:anomaly} Anomaly]{\includegraphics[width=0.47\linewidth]{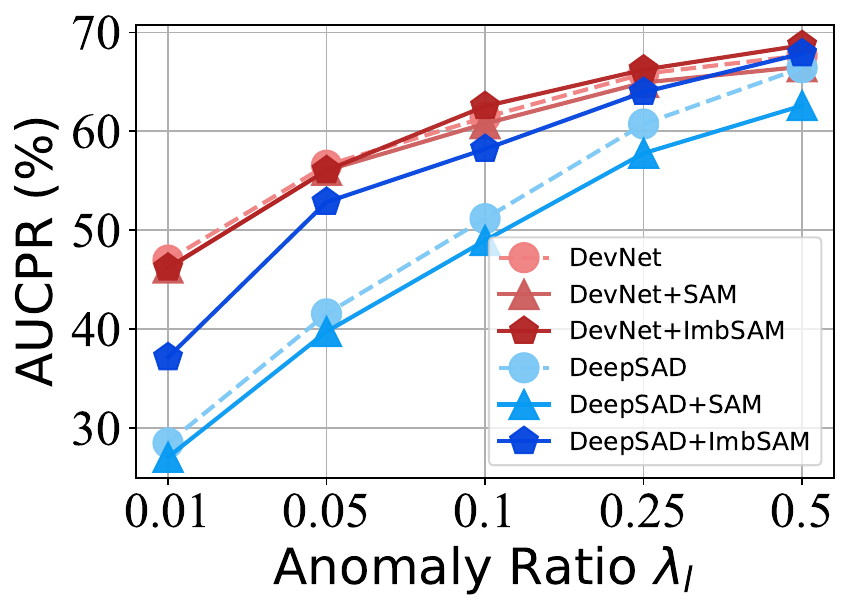}}
   \hspace{0.02\linewidth}
   \subfigure[\label{fig:abla-ssad:normal} Normal]{\includegraphics[width=0.47\linewidth]{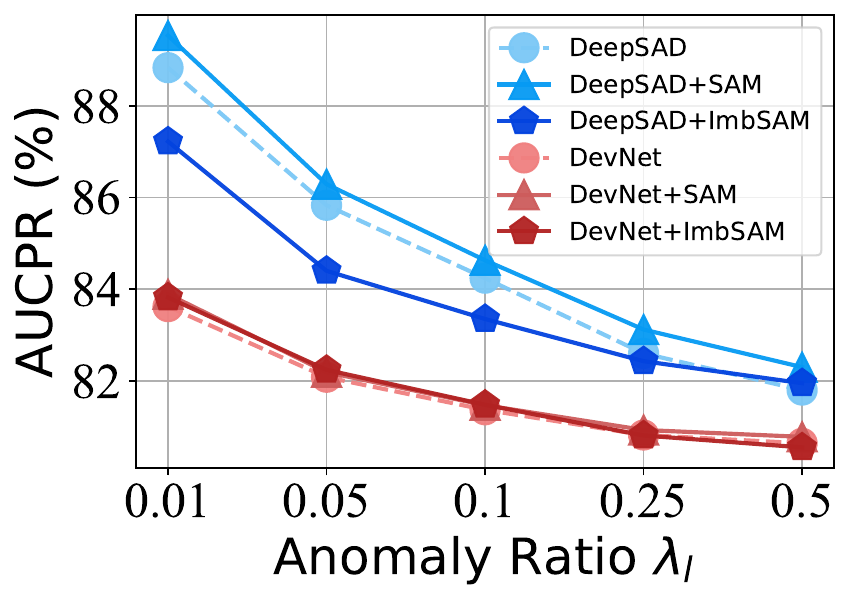}}
   \caption{\tbf{AUCPR of anomaly and normal} for the DeepSAD \cite{ruff2019deep} and DevNet \cite{pang2019devnet} optimized by SGD \cite{bottou2012stochastic}, SAM \cite{foret2020sharpness} and our ImbSAM, averaged on five SSAD datasets.}
   \label{fig:abla-ssad}
\end{figure}

To further confirm the generalization scope of our ImbSAM, we report the AUCPR for normal and anomaly respectively, with an increasing anomaly ratio $\gamma_l\in$ [1\%, 5\%, 10\%, 25\%, 50\%]. Notably, the imbalanced factor $IF \gg 1$ even when $\gamma_l = 50\%$. As displayed in Figure \ref{fig:abla-ssad}, our ImbSAM significantly improves the AUCPR of anomaly for DeepSAD by about 10\%, comparable with the DevNet that introduces re-sampling (Figure \ref{fig:abla-ssad:anomaly}), while SAM only slightly increases AUCPR on the side of normal (Figure \ref{fig:abla-ssad:normal}). In particular, with $\gamma_l<$0.1, the anomaly AUCPR of DevNet slightly drops when optimized with our ImbSAM. This also verifies our analysis in Section \ref{sec:sam-limitation} that overexpose of limited data conflicts with SAM.

\section{Conclusion}
In order to adapt the promising SAM to tackle the overfitting issues in class-imbalanced recognition, we leverage class priors to control the generalization scope of SAM to focus tail classes and propose a class-aware ImbSAM. Our ImbSAM demonstrates remarkable performance improvement especially for the classes with limited training data, achieving novel SOTA in both long-tailed classification and semi-supervised anomaly detection.

\section*{Acknowledgments}
This work was supported in part by the National Natural Science Foundation of China (62222203, 61976049) and CAAI-Huawei MindSpore Open Fund.

\newpage\newpage
{\small
   \bibliographystyle{ieee_fullname}
   \bibliography{egbib_zyx}
}

\end{document}